\documentclass[conference]{IEEEtran}
\IEEEoverridecommandlockouts
\usepackage{cite}
\usepackage{amsmath,amssymb,amsfonts}
\usepackage{algorithmic}
\usepackage{graphicx}
\usepackage{textcomp}
\usepackage{xcolor}
\def\BibTeX{{\rm B\kern-.05em{\sc i\kern-.025em b}\kern-.08em
    T\kern-.1667em\lower.7ex\hbox{E}\kern-.125emX}}

\usepackage{latexsym}
\usepackage{amssymb}
\usepackage{amsmath}
\usepackage{amsthm}
\usepackage{multirow}
\usepackage{url}
\usepackage{graphicx}
\usepackage{color}

\begin{document}

\title{Collecting Indicators of Compromise \\ 
from Unstructured Text of Cybersecurity Articles \\ 
using Neural-Based Sequence Labelling}

\author{\IEEEauthorblockN{
\begin{tabular}{ccc}
Zi Long & Lianzhi Tan & Shengping Zhou
\end{tabular}}
\IEEEauthorblockA{\textit{Platform and Content Group} \\
  \textit{Tencent}\\
  Shenzhen, Guangdong, China}
\and
\IEEEauthorblockN{Chaoyang He}
\IEEEauthorblockA{\textit{AI Lab} \\
  \textit{Tencent}\\
  Shenzhen, Guangdong, China}
\and
\IEEEauthorblockN{Xin Liu}
\IEEEauthorblockA{\textit{Platform and Content Group} \\
  \textit{Tencent}\\
  Shenzhen, Guangdong, China}
}

\maketitle

\begin{abstract}
  Indicators of Compromise (IOCs) are artifacts observed on a network or in an operating system that can be utilized to indicate a computer intrusion and detect cyber-attacks in an early stage. Thus, they exert an important role in the field of cybersecurity. However, state-of-the-art IOCs detection systems rely heavily on hand-crafted features with expert knowledge of cybersecurity, and require large-scale manually annotated corpora to train an IOC classifier. In this paper, we propose using an end-to-end neural-based sequence labelling model to identify IOCs automatically from  cybersecurity articles without expert knowledge of cybersecurity. 
%   Our work is the first to apply an end-to-end sequence labelling to the task in IOCs identification. 
  By using a multi-head self-attention module and contextual features, we find that the proposed model is capable of gathering contextual information from texts of cybersecurity articles and performs better in the task of IOC identification.
  Experiments show that the proposed model outperforms other sequence labelling models, achieving the average F1-score of 89.0\% on English cybersecurity article test set, and approximately the average F1-score of 81.8\% on Chinese test set.
\end{abstract}

% \begin{IEEEkeywords}
% component, formatting, style, styling, insert
% \end{IEEEkeywords}

\section{Introduction}
\label{sec:intro}

\begin{figure}[tb]
\centering
\caption{Example of IOCs contained in cybersecurity articles}
\includegraphics[scale=0.4]{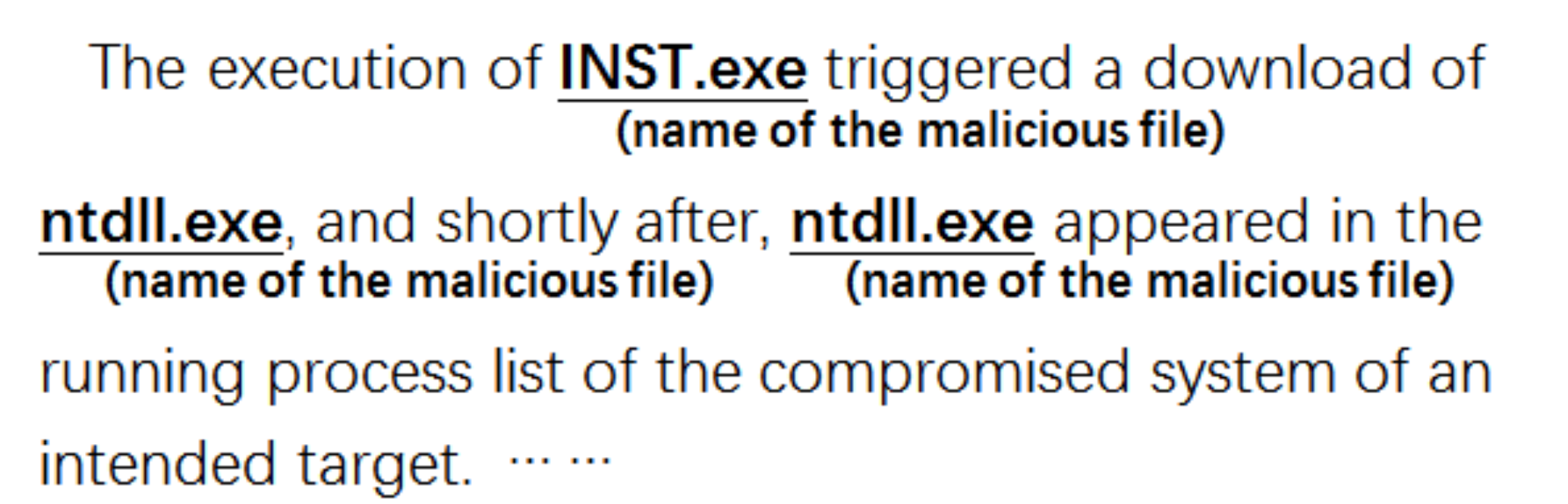}
\label{fig:ioc_example}
\end{figure}

 Indicators of Compromise (IOCs) are forensic artifacts that are used as signs when a system has been compromised by an attacker or infected with a particular piece of malware. To be specific, IOCs are composed of some combinations of virus signatures, IPs, URLs or domain names of botnets, MD5 hashes of attack files, etc. They are frequently described in cybersecurity articles, many of which are written in unstructured text, describing attack tactics, technique and procedures.
 For example,
 a snippet from a cybersecurity article is shown in Fig.~\ref{fig:ioc_example}. From the text , token ``INST.exe'' is the name of an executable file of a malicious software, and the file ``ntdll.exe'' downloaded by ``INST.exe'' is a malicious file as well. Obviously, these kinds of IOCs can be then utilized for early detection of future attack attempts by using intrusion detection systems and antivirus software, and thus, they exert an important role in the field of cybersecurity.
 However, with the rapid evolvement of cyber threats, the IOC data are produced at a high volume and velocity every day, which makes it increasingly hard for human to gather and manage them. 
 
 A number of systems are proposed to help discover and gather malicious information and IOCs from various types of data  sources\cite{ZhuZ16,LiaoX16,Husari17,HuangC17,Kwon17,ZhuZ18}. 
 However, most of those systems consist of several components that identify IOCs by using human-crafted features that heavily rely on specific language knowledge such as dependency structure, and they often have to be pre-defined by experts in the field of the cybersecurity. Furthermore, they need a large amount of annotated data used as the training data to train an IOC classifier. Those training data are frequently difficult to be crowed-sourced, because non-experts can hardly distinguish IOCs from those non-malicious IPs or URLs. Thus, it is a time-consuming and laborious task to construct such systems for different languages.
%  a system.
 
%%%%%%%%%%%%%%%%%%%
%%%%%%%%%%%%%%%%%%

 In this work, we consider the task of collecting IOCs from cybersecurity articles as a task of sequence labelling of natural language processing (NLP). By applying a sequence labelling model, each token in an unstructured input text is assigned with a label, and tokens assigned with IOC labels are then collected as IOCs.
 Recently, sequence labelling models have been utilized in many NLP tasks.
 Huang et al.\cite{HuangZ15} proposed using a sequence labelling model based on the bidirectional long short-term memory (LSTM)\cite{Hochreiter97} for the task of named entity recognition (NER). Chiu et al.\cite{Chiu16} and Lample et al.\cite{Lample16} proposed integrating LSTM encoders with character embedding and the neural sequence labelling model to achieve a remarkable performance on the task of NER as well as part-of-speech (POS) tagging. Besides, Dernoncourt et al.\cite{Dernoncourt17} and Jiang et al.\cite{JiangZ17} proposed applying the neural sequence labelling model to the task of de-identification of medical records. 
%  To the best of our knowledge, we are the first to apply an end-to-end sequence labelling to the task of IOCs identification in the field of cybersecurity.
 %
 
 Among the previous studies of the neural sequence labelling task, 
 Zhou el al.\cite{Zhou18} firstly propose using an end-to-end neural sequence labelling model to fully automate the process of IOCs identification. 
 Their model is on the basis of an artificial neural networks (ANN) with bidirectional LSTM and CRF. 
 However, their newly introduced spelling features bring a more extraction of false positives, i.e., tokens that are similar to IOCs but not malicious.  
 In this paper, we further introduce a multi-head self-attention module and contextual features to the ANN model so that the proposed model can perform better in gathering the contextual information from the unstructured text for the task of IOCs identification. 
 Based on the results of our experiments, our proposed approach achieves an average precision of 93.1\% and the recall of 85.2\% on English cybersecurity article test set, and an average precision of 82.9\% and recall of 80.7\% on Chinese test set. 
 We further evaluate the proposed model by training the model using both the English dataset and Chinese dataset, which even achieves better performance.

%%%%%%%%%%%%%%%%%%%
\begin{figure*}[tb]
\centering
\includegraphics[scale=0.36]{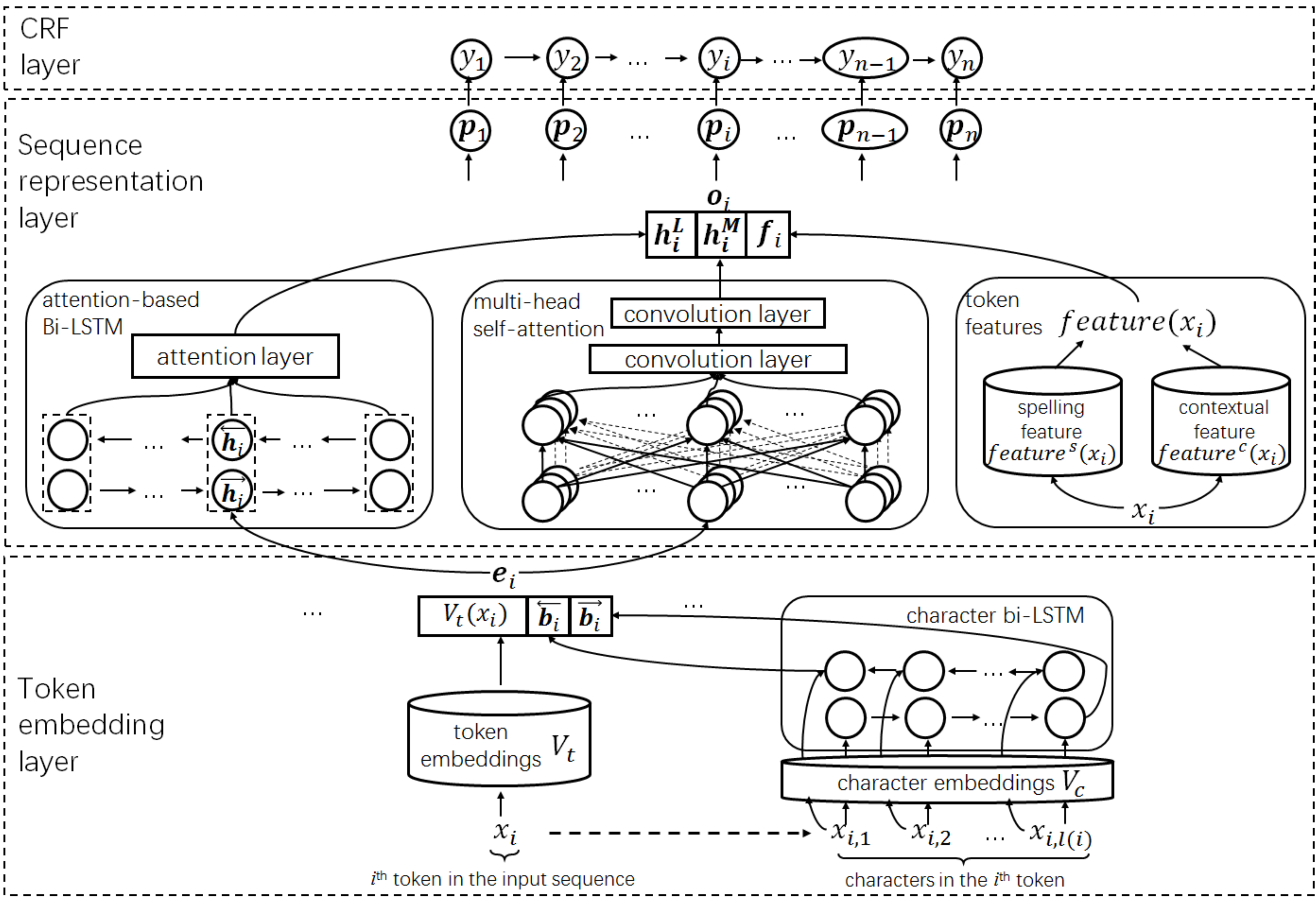}
\caption{ANN model of sequence labeling for IOCs automatic identification}
\label{fig:model}
\end{figure*}
%%%%%%%%%%%%%%%%%%

\section{Model}
\label{sec:model}

 Fig.~\ref{fig:model} shows the 3 components (layers) of the proposed neural network architecture.

\subsection{Token Embedding Layer}
\label{sec:embedding}

 The token embedding layer takes a token as input and outputs its vector representation.
 As shown in Fig.~\ref{fig:model}, given an input sequence of tokens $x_{1}, \ldots, x_{n}$, the output vector $\boldsymbol{e}_{i}$ ($i =1,\ldots,n$) of each token $x_{i}$ results from the concatenation of two different types of embeddings: token embedding $V_t(x_i)$ and the character-based token embeddings $\overrightarrow{b_i}$, $\overleftarrow{b_i}$ that come from the output of a character-level bi-LSTM encoder. 
%  The output vector results from the concatenation of two different types of embeddings: the first one directly maps a token to a vector, while the second one outputs a character-level token encoder.
 
%  As shown in Fig.~\ref{fig:model}, given an input sequence of tokens $x_{1}, \ldots, x_{n}$, each token $x_{i}$ ($i = 1, \ldots, n$) is mapped to a token embedding $V_t(x_i)$ with the mapping of token embedding $V_t(\cdot)$. The token embedding is pre-trained on large unlabeled datasets, and then learned jointly with the rest of the model. Then, let $x_{i,1}, \ldots, x_{i,l(i)}$ be the sequence of characters that comprise the token $x_{i}$, where $l(i)$ is the number of characters in $x_{i}$. Each character $x_{i,j}$ ($j=1, \ldots, l(i)$) is mapped to a character embedding $V_c(x_{i,j})$ using the mapping of character embedding $V_c(\cdot)$. 
%  The character embedding is randomly initialized and also jointly learned during the training process.
%  Then the vector $V_c(x_{i,j})$ is passed to a bidirectional LSTM, which outputs a forward character-based token embedding $\overrightarrow{b_i}$ and a backward embedding $\overleftarrow{b_i}$. 
%  Finally, the output $\boldsymbol{e}_{i}$ of the input embedding layer for the $i^{th}$ token $x_{i}$ is the concatenation of the token embedding $V_t(x_i)$ and the character-based token embeddings $\overrightarrow{b_i}$, $\overleftarrow{b_i}$.

\subsection{Sequence Representation Layer}
\label{sec:lstm}

 The Sequence Representation Layer takes the sequence of embeddings $\boldsymbol{e}_{i}$ ($i =1,\ldots,n$) as input, and outputs a sequence $\boldsymbol{p}_{i}(i =1,\ldots,n)$, where the $t^{th}$ element of $\boldsymbol{p}_i$ represents the probability that the $i^{th}$ token has the label $t$.
 
 Different from the previous work of sequence labelling in news articles or patient notes\cite{Lample16,Dernoncourt17}, sentences from a cybersecurity report often contain a large number of tokens as well as lists of IOCs with little context, making it much more difficult for LSTM to encode the input sentence correctly. 
 Therefore, instead of the token LSTM layer in \cite{Zhou18}, we propose sequence representation layer that consists of 3 modules, i.e., attention-based Bi-LSTM module, multi-head self-attention module and token feature module. 

\paragraph{Attention-based Bi-LSTM}
 Considering that tokens cannot contribute equally to the representation of the input sequence, we introduce attention mechanism to Bi-LSTM to extract such tokens that are crucial to the meaning of the sentence. Then, we aggregate the representation of those informative words to form the vector of the input sequence. The attention mechanism is similar to the one proposed by Yang et al.\cite{Yang16}, which is defined as follows:
 \begin{eqnarray}
  \boldsymbol{u}_{i} & = & \tanh(W_{w}\boldsymbol{h}_{i} + \boldsymbol{b}_{w}) \nonumber \\
  \alpha_{i} & = & \frac{\exp(\boldsymbol{u}_{i}^\top\boldsymbol{u}_{w})}{\sum_{i}\exp(\boldsymbol{u}_{i}^\top\boldsymbol{u}_{w})} \nonumber \\
  \boldsymbol{s} & = & \sum_{i} \alpha_{i}\boldsymbol{h}_{i} \nonumber
 \end{eqnarray}
 That is to say, we first compute the $\boldsymbol{u}_i$ as a hidden representation of the hidden states of Bi-LSTM $\boldsymbol{h}_{i}$ for $i^{th}$ input token, where $\boldsymbol{h}_{i}$ is obtained by concatenating the $i^{th}$ hidden states of forward and backward LSTM, i.e., $\boldsymbol{h}_{i} = [\overrightarrow{\boldsymbol{h}}_i; \overleftarrow{\boldsymbol{h}}_i]$.
 Then, we measure the importance of the $i^{th}$ token with a trainable vector $\boldsymbol{u}_w$
 and get a normalized importance weight $\alpha_{i}$ through a softmax function. After that, the sentence vector $\boldsymbol{s}$ is computed as a weighted sum of $\boldsymbol{h}_{i}$ ($i=1,\ldots,n$). Here, weight matrix $W_{w}$, bias $\boldsymbol{b}_{w}$ and vector $\boldsymbol{u}_w$ are randomly initialized and jointly learned during the training process. Note that each input sentence merely has one sentence vector $\boldsymbol{s}$ as its weighted representation, and $\boldsymbol{s}$ is then used as a part of the $i^{th}$ output of attention-based Bi-LSTM module, where $\boldsymbol{h}^{L}_{i} = [\overrightarrow{\boldsymbol{h}}_i; \overleftarrow{\boldsymbol{h}}_i; \boldsymbol{s}]$ ($i=1,\ldots,n$).
 
 \paragraph{Multi-head self-attention}
 Motivated by the successful application of self-attention in many NLP tasks\cite{Vaswani17,XieJ18}, we add a multi-head self-attention module to enhance the embedding of each word with the information of other words in a text adaptively. By means of this, the local text regions where convolution performs carry the global information of text. Following the encoder part of Vaswani et al.\cite{Vaswani17}, multi-head self-attention module is composed of a stack of several identical layers, each of which consists of a multi-head self-attention mechanism and two convolutions with kernel size 1. Given the sequence of embeddings $\boldsymbol{e}_{i}$ as input, and the output is defined as follows:
  \begin{eqnarray}
     \textup{Attention}(Q,K,V) & = & \textup{softmax}(\frac{QK^{T}}{\sqrt{d_{k}}}) \nonumber\\ 
     \textup{MultiHead}(Q,K,V) & = & [\textup{head}_1;\ldots;\textup{head}_h] \nonumber \\ 
     where\ \textup{head}_i & = & \textup{Attention}(QW^Q_i,KW^K_i,VW^V_i) \nonumber
  \end{eqnarray}
   where, $W^Q_i$, $W^K_i$, $W^V_i$ are parameter matrices for the projections of queries $Q$, keys $K$ and values $V$ in the $i^{th}$ head, respectively. 
   Here, $Q$, $K$ and $V$ are set as the input sequence $\boldsymbol{e}_{i}$ ($i=1,\ldots,n$).
%   , and dimension of keys $d_{k}$ is the same as the dimension of $\boldsymbol{e}_{i}$. 
   The $\textup{MultiHead}(Q,K,V)$ is then given to the two convolutions and the output of multi-head self-attention $\boldsymbol{h}^{M}_{i}$ ($i=1,\ldots,n$) is obtained.

 \paragraph{Token features}
 Furthermore, we introduce some features to defined IOCs to improve the performance of the proposed model on a very small amount of training data. Here, we define two types of features, i.e., spelling features and contextual features, and map each token $x_i$ ($x=1, \ldots, n$) to a feature vector $feature(x_i) = [feature^{s}(x_i);feature^{c}(x_i)]$, where $feature^{s}(x_i)$ is the spelling feature vector and $feature^{c}(x_i)$ is the contextual feature vector. 
%  The $q^{th}$ element of $feature(x_i)$ represents the value of the $q^{th}$ feature of token $x_i$. 
 Note that the values of features are then jointly learned during the process of training. In Section~\ref{sec:feat}, we will explain the features in more detail.
 
 As shown in Fig.~\ref{fig:model}, the vector $\boldsymbol{o}_{i}$ ($i=1,\ldots,n$) is a concatenation of the $\boldsymbol{h}^{L}_{i}$, $\boldsymbol{h}^{M}_{i}$ and $f(i)$. Each vector $\boldsymbol{o}_{i}$ is then given to a feed-forward neural network with one hidden layer, which outputs the corresponding probability vector $\boldsymbol{p}_{i}$.

\subsection{CRF Layer}
\label{sec:predict}

 We also introduce a CRF layer to output the most likely sequence of predicted labels. 
 The score of a label sequence $\boldsymbol{y} = y_1, \ldots, y_{n}$ is defined as the sum of the probabilities of unigram labels and the bigram label transition probabilities:
 \begin{eqnarray}
     score(\boldsymbol{y}) = \sum^n_{i=1} p_i[y_i] + \sum^n_{i=2} T[y_{i-1}, y_{i}] \nonumber
 \end{eqnarray}
 where $T$ is a matrix that contains the transition probabilities of two subsequent labels. Vector $\boldsymbol{p}_{i}$ is the output of the token LSTM layer, and $\boldsymbol{p}_{i}[y_i]$ is the probability of label $y_i$ in $\boldsymbol{p}_{i}$. $T[g,h]$ is the probability that a token with label $g$ is followed by a token with the label $h$. Subsequently, these scores are turned into probabilities of the label sequence by taking a softmax function over all possible label sequences. %During training, the objective it to maximize the log probability of the correct label sequences. While decoding, given an input sequence of tokens, the corresponding sequence of predicted labels is chosen as the one that maximize the score.

\section{Features}
\label{sec:feat}
 We extract a vector of features for each tokens of input sequences.
%  , which we then concatenate it with the output of attention-based Bi-LSTM module and multi-head self-attention module
 In this section, we present each feature category in detail.
 
\subsection{Spelling Features}
\label{sec:feature}
 
 Since the IOCs tend to follow fixed patterns, we predefined several regular expressions and spelling rules to identify IOCs. For example, to identify a URL, we defined a regular expression $\verb|http(s)?:\\[0-9a-zA-Z_\.\-\\]+|$ and set the value of the URL feature to 1 when the input token matches the regular expression.\footnote{ 
   Details of the spelling features can be found in \cite{Zhou18}.
 }
 However, such expressions and spelling rules could introduce false positives, i.e., tokens that have the same spelling patterns as IOCs but are not malicious. In this work, we further introduce the contextual features as described next.

\subsection{Contextual Features}
\label{sec:context_feature}
 IOCs in cybersecurity articles are often described in a predictable way: being connected to a set of contextual keywords\cite{Darling15,LiaoX16}. 
 For example, a human user can infer that the word ``ntdll.exe'' is the name of a malicious file on the basis of the words ``download'' and ``compromised'' from the text shown in Fig.~\ref{fig:ioc_example}. 
 By analyzing the whole corpus, it is interesting that malicious file names tends to co-occur with words such as "download", "malware", "malicious", etc. 
 In this work, we consider words that can indicate the characteristics of the neighbor words as contextual keywords and develop an approach to generate features from the automatically extracted contextual keywords.
 
%  Given an input token $x$, the $i^{th}$ element of contextual feature vector$feature^{c}(x)$ is defined as follows:
 Taking the above into account, we introduce the contextual feature vector $feature^{c}(x)$ for a given input token $x$, where the $i^{th}$ element of $feature^{c}(x)$ is defined as follows:
 \begin{eqnarray}
   feature_i^{c}(x) & = & \frac{freq_{w}(k_{i};x,size)}{freq(x)} \nonumber \\
     & & (k_{i} \in K, i = 1,\ldots,|K|) \nonumber 
 \end{eqnarray}
 $freq(x)$ is the frequency of token $x$ in the whole corpus, while $freq_{w}(k_{i};x,size)$ is the frequency of contextual keyword $k_{i}$ from the windowed portions of the texts centering on the token $x$ in the whole corpus and $size$ is the size of window. The set of contextual keywords $K$ are automatically extracted from the annotated texts, where each contextual keyword $k_i$($k_i \in K$) satisfies the following conditions: 
 \begin{description}
   \item[(1)] $\sum\limits_{t \in T} freq_{w}(k_{i};t,size) \geq lb$, where $T$ is the set of manually annotated IOCs and $lb$ is a the lower bound of the frequency.
   \item[(2)] $k_{i}$ is not a punctuation or stopword\footnote{
     \url{https://github.com/hankcs/HanLP/blob/master/data/dictionary/stopwords.txt} 
   }. 
 \end{description}
 
 Note that we extract contextual keywords only from manually annotated data (e.g., training set), while we compute the contextual feature vector in all of the unlabeled data.
 According to this definition, it is obvious that the dimension of the contextual feature vector is as the same as the number of extracted contextual keywords. The size of window $size$ and the lower bound of frequency $lb$ are then tuned by the validation set.

\subsection{Usage of Features}
\label{sec:feature_usage}

 The feature vector for an input token is the concatenation of the token spelling feature vector and the contextaul feature vector.
 %and use it Two feature vectors are concatenated and used as a vector.
%  , and are concatenated with the LSTM hidden state vector and the sentence vector of attention in the token LSTM layer as shown in Section~\ref{sec:lstm}.
 Here, to elucidate the best usage of the feature vector, 
 we evaluate the feature vector by concatenating it at different locations in the proposed model, i.e., the input of the token LSTM layer ($e_{i} = [V_t(x_i);\overrightarrow{\boldsymbol{b}_i};\overleftarrow{\boldsymbol{b}_i};feature(x_i)]$), the hidden state of the token LSTM ($\boldsymbol{h}_{i} = [\overrightarrow{\boldsymbol{h}}_i;\overleftarrow{\boldsymbol{h}}_i;feature(x_i)]$), and the output of token LSTM ($\boldsymbol{o}_i=[\boldsymbol{h}_{i};\boldsymbol{s};feature(x_i)]$).
 Among them, to concatenate the feature vector with the LSTM hidden state vector and the sentence vector of attention in the token LSTM layer, as shown in Section~\ref{sec:lstm}, achieved the best performance.
 We speculate that the features played an important role in the task of IOCs identification and feature vectors near the output layer were able to improve the performance more significantly than those at other locations.

%%%%%%%%%%%%%%%%%%%%%%%%%%%%%%%%%%%%%%%%%%%
\begin{table}
 \centering
 \caption{Statistics of datasets (Numbers of training / validation / test set)}
 \label{tab:dataset}
 \begin{tabular}{|p{2cm}||c||c|}
    \hline
     & English Dataset & Chinese Dataset \\ \hline\hline
     %& set & set & \\     
    attacker & 5,304 / 1,067 / 1,609 & 742 / 230 / 126 \\ \hline
    attack method & 2,737 / 610 / 882 & 126 / 41 / 27 \\ \hline
    attack target & 3,055 / 1,055 / 695 & 161 / 17 / 15 \\ \hline
    domain & 6,443 / 1,054 / 1,701 & 1,682 / 438 / 468 \\ \hline
    e-mail address & 1,284 / 154 / 222 & 67 / 16 / 49 \\ \hline
    file hash & 10,367 / 2,055 / 2,459 & 2,484 / 1,231 / 1,055 \\ \hline
    file information & 4,353 / 1,024 / 1,131 & 931 / 469 / 531 \\ \hline
    IPv4 & 3,012 / 729 / 819 & 969 / 252 / 264 \\ \hline
    malware & 7,317 / 1,585 / 1,974 & 2,958 / 1,524 / 1,035 \\ \hline
    URL & 1,849 / 105 / 156 & 1,618 / 333 / 487 \\ \hline
    vulnerability & 1,557 / 309 / 359 & 469 / 166 / 143 \\ \hline \hline
    tokens & 1,169,896 / 253,336  & 775,549 / 282,600 \\
     &  / 350,406 &  / 249,946 \\ \hline
    paragraphs & 6,702 / 1,453 / 2,110 & 6,423 / 2,314 / 2,059\\ \hline
    articles & 250 / 70 / 70 & 363 / 122 / 122 \\ \hline
 \end{tabular}
\end{table}
%%%%%%%%%%%%%%%%%%%%%%%%%%%%%%%%%%%%%%%%%%%%

\section{Evaluation}
\label{sec:eva}

\subsection{Datasets}
\label{sec:data}

 For English dataset, we crawl 687 cybersecurity articles from a collection of advanced persistent threats (APT) reports which are published from 2008 to 2018\footnote{
    \url{https://github.com/CyberMonitor/APT_CyberCriminal_Campagin_Collections}
 }. All of these cybersecurity articles are used to train the English word embedding. 
 Afterwards, we randomly select 370 articles, and manually annotate the IOCs contained in the articles. Among the selected articles, we randomly select 70 articles as the validation set and 70 articles as the test set; the remaining articles are used for  training. 
 
 For Chinese dataset, we crawl 5,427 cybersecurity articles online from 35 cybersecurity blogs which are published from 2001 to 2018. All of these cybersecurity articles are used to train the Chinese word embedding. Afterwards, we randomly select 607 articles, and manually annotate the IOCs contained in the articles. Among the selected articles, we randomly select 122 articles as the validation set and 122 articles as the test set; the remaining articles are used for training. 
 
 TABLE~\ref{tab:dataset} shows statistics of the datasets. The output labels are annotated with the BIO (which stands for ``Begin'', ``Inside'' and ``Outside'') scheme.

\subsection{Training Details}
\label{sec:train}
 
 For pre-trained token embedding, we apply word2vec\cite{Mikolov13} to all crawled 687 English APT reports and 5,427 Chinese cybersecurity articles described in Section~\ref{sec:data} respectively. The word2vec models are trained with a window size of 8, a minimum vocabulary count of 1, and 15 iterations. The negative sampling number of word2vec is set to 8 and the model type is skip-gram. The dimension of the output token embedding is set to 100.
 
 The ANN model is trained with the stochastic gradient descent to update all parameters, i.e., token embedding, character embedding, parameters of Bi-LSTM, weights of sentence attention, weights of multi-head self-attention, token features, and transition probabilities of CRF layers at each gradient step. 
%  For regularization, the dropout is applied to the character-enhanced token embedding before the token LSTM layer. 
 For regularization, the dropout is applied to the output of each sub layer of the ANN model.
 Further training details are given below: 
  (a) For attention-based Bi-LSTM module, dimensions of character embedding, hidden states of character-based token embedding LSTM, hidden states of Bi-LSTM, and sentence attention are set to 25, 25, 100 and 100, respectively. For multi-head self-attention module, we employ a stack of 6 multi-head self attention layer, each of which has 4 head and dimension of each head is set to 64. 
 (b) All of the ANN’s parameters are initialized with a uniform distribution ranging from -1 to 1.
 (c) We train our model with a fixed learning rate of 0.005. The minimum number of epochs for training is set as 30. After the first 30 epochs had been trained, we compute the average F1-score of the validation set by the use of the currently produced model after every epoch had been trained, and stop the training process when the average F1-score of validation set fails to increase during the last ten epochs. 
 We train our model for, if we do not early stop the training process, 100 epochs as the maximum number. 
 (d) We rescale the normalized gradient to ensure that its norm does not exceed 5. 
 (e) The dropout probability is set to 0.5. 
 
%  We train the ANN model on the training set. The training time is around 20 hours when using the described parameters on an 8-CPU machine.
%%%%%%%%%%%%%%%%%%%%%%%%%%%%%%%%%%%%%%%%%%%%%%
\begin{table}[t]
 \centering
 \caption{evaluation results (micro average for 11 labels) }
 \label{tab:result}
 \begin{tabular}{|p{1.8cm}||c|c|c||c|c|c|}
    \hline
    \!\!\!Models\!\!\!& \multicolumn{3}{c||}{English Dataset} & \multicolumn{3}{c|}{Chinese Dataset} \\ \cline{2-7}
    \!\!\! \!\!\! &\!\!\!Precision\!\!\!&\!\!\!Recall\!\!\!&\!\!\!F1-score\!\!\!&\!\!\!Precision\!\!\!&\!\!\!Recall\!\!\!&\!\!\!F1-score\!\!\!\\ \hline \hline
    \!\!\!Baseline\!\!\!& 47.1 & 58.8 & 52.3 & 60.1 & 42.7 & 49.9 \\ \hline\hline
    %CRF\cite{Lafferty01} & 19.3 & 7.8 & 8.8 \\ \hline
    \!\!\!Huang et al.\cite{HuangZ15}\!\!\!& 64.8 & 33.6 & 51.6 & 60.2 & 35.0 & 44.3 \\ \hline
    \!\!\!Lample et al.\cite{Lample16}\!\!\!& 83.0 & 75.2 & 78.9 & 78.3 & 71.8 & 74.9 \\ \hline
    \!\!\!Rei et al.\cite{ReiM16}\!\!\!& 81.6 & 74.5 & 77.9 & 76.1 & 71.0 & 73.5 \\ \hline \hline
    %the proposed model (Char-LSTM-CRF models with attention mechanism) & {\bf 90.4} & 73.3 & 81.0 \\ \hline
    \!\!\!Zhou et al.\cite{Zhou18}\!\!\!& 90.4 & {\bf 87.2} & 88.8 & 80.8 & 80.1 & 80.4 \\ \hline \hline
    \!\!\!Our model\!\!\!& {\bf 93.1} & {\bf 85.2} & {\bf 89.0} & {\bf 82.9} & {\bf 80.7} & {\bf 81.8} \\ \hline
 \end{tabular}
\end{table}
%%%%%%%%%%%%%%%%%%%%%%%%%%%%%%%%%%%%%%%%%%%%%
\subsection{Results}
\label{sec:result}

 As shown in TABLE~\ref{tab:result}, we report the micro average of precision, recall and F1-score for all 11 types of labels for a baseline as well as the proposed model.
 As the baseline, we simply judge the input token as IOCs on the basis of the spelling features described in \cite{Zhou18}. As presented in TABLE~\ref{tab:result}, the score obtained by the proposed model is clearly higher than the baseline.
 Here, as described in Section~\ref{sec:context_feature}, the sizes of window and lower bounds of frequency for selecting contextual keywords are tuned as 4 and 7 throughout the evaluation of English dataset, and tuned as 3 and 4 throughout the evaluation of Chinese dataset. The number of extracted contextual keywords from  the English dataset is 1,328, and from the Chinese dataset is 331. 

% %%%%%%%%%%%%%%%%%%%
% \begin{figure}[ht]
% \caption{Impact of the training set size on F1-score}
% \label{fig:data_size}
% \centering
% \includegraphics[scale=0.53]{figures/train_data_size.eps}
% \end{figure}
% %%%%%%%%%%%%%%%%%%
%%%%%%%%%%%%%%%%%%%
\begin{table*}[!t]
\caption{Examples of correct identification by the proposed model }
\label{tab:correct_example}
\centering
\includegraphics[scale=0.36]{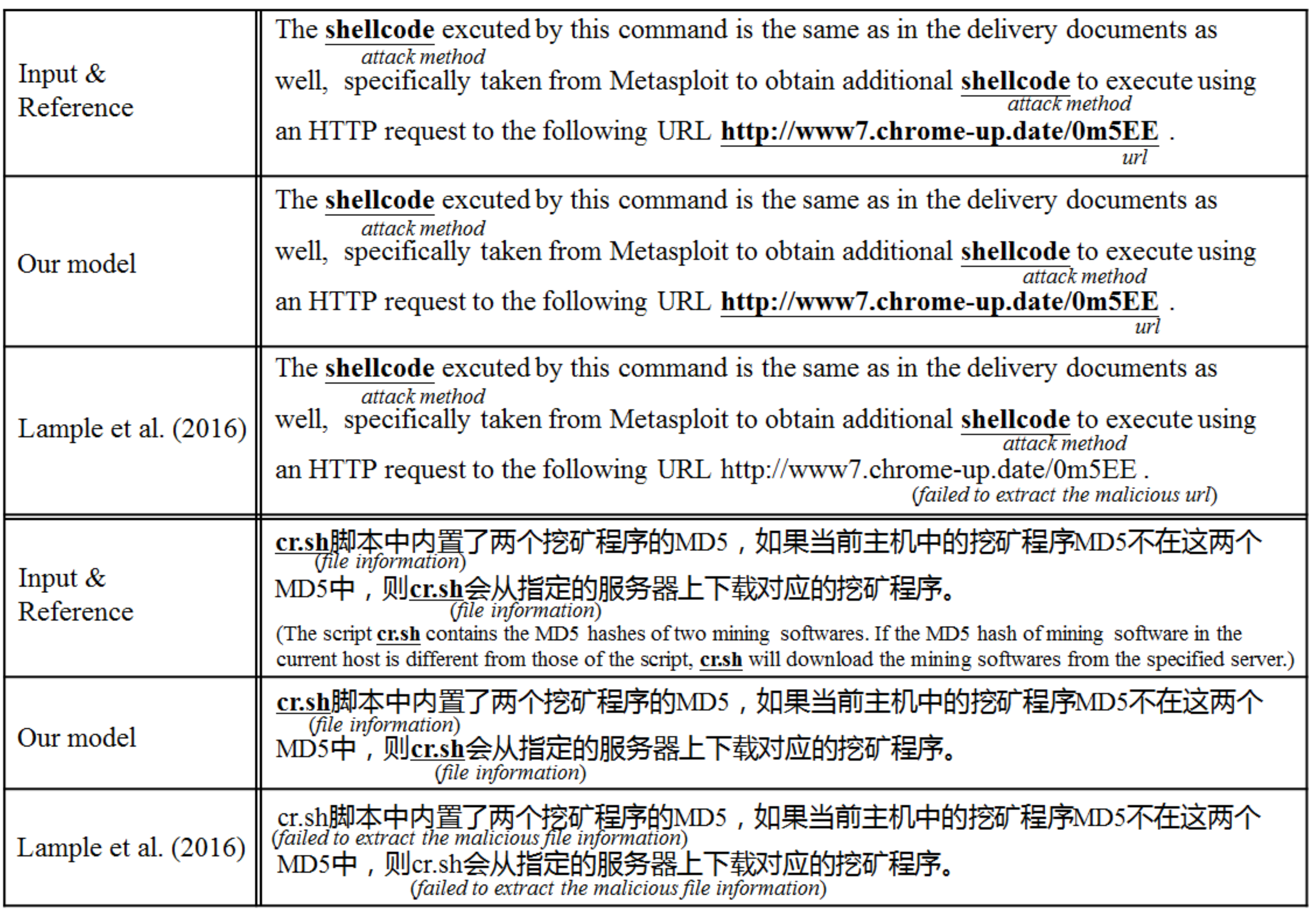}
\end{table*}
%%%%%%%%%%%%%%%%%%

 Furthermore, we quantitatively compare our study with other typical works of sequence labelling, i.e., the work of Huang et al.\cite{HuangZ15}, the work of Lample et al.\cite{Lample16} and the work of Rei et al.\cite{ReiM16}. 
 Huang et al.\cite{HuangZ15} proposed a bidirectional LSTM model with a CRF layer, including hand-crafted features specialized for the task of sequence labelling. Lample et al.\cite{Lample16} described a model where the character-level representation was concatenated with word embedding and Rei et al.\cite{ReiM16} improved the model by introducing an attention mechanism to the character-level representations.
 We train these models by employing the same training set and training parameters as the proposed model. As shown in TABLE~\ref{tab:result}, the proposed model obtains the highest precision, recall and F1-score than other models in the task of IOCs extraction. Compared with the second-best model of Lample et al.\cite{Lample16}, the performance gain of the proposed model on the English dataset is approximately 10.1\% of precision and 10.0\% of recall. The performance gain of the proposed model on the Chinese dataset is approximately 4.2\% of precision and 9.0\% of recall.

 We also quantitatively compare our study with the work of Zhou et al.\cite{Zhou18}, which proposed a bidirectional LSTM model with a CRF layer, including hand-crafted spelling features for the task of IOC identification. As shown in TABLE \ref{tab:result}, the proposed model obtains a slightly higher F1-score on the English dataset and significantly higher F1-score on the Chinese dataset.
 
 TABLE~\ref{tab:correct_example} compares several examples of correct IOC extraction produced by the proposed model with one by the work of Lample et al.\cite{Lample16}. In the first example, the model of Lample et al.\cite{Lample16} fails to identify the malicious URL ``http://www7.chrome-up.date/0m5EE'', because the token only appears in the test set and consists of several parts that are uncommon for URLs, such as ``www7'' and ``date'', and thus both the token embedding and the character embedding lack proper information to represent the token as a malicious URL. The proposed model correctly identifies the URL, where the token is defined as a URL by spelling features and is then identified as a malicious URL by the use of the context information. 
%  In the second example, token ``MDDEFGEGETGIZ'' is erroneously identified as the name of a malicious file by the model of Lample et al.\cite{Lample16} because of the context ``DLL's config'' before the token that tends to co-occur with names of files. The token is correctly identified by the proposed model, because the token fails to match the regular expression of file information, and is consequently not considered as a name of a malicious file.
 In the second example, the model of Lample et al.\cite{Lample16} fails to identify token ``cr.sh'' of the input Chinese text as a malicious file name, while the token is assigned with a correct label by the proposed model. It is mainly because that the token ``cr.sh'' is defined as a token of file information by spelling features and tends to co-occur with words, ``\includegraphics[scale=0.35]{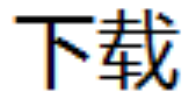}''(download) and ``\includegraphics[scale=0.35]{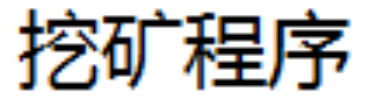}''(mining software).
 These two words often appear nearby malicious file information and are then extracted as contextual keywords in Section~\ref{sec:context_feature}. The token ``cr.sh'' is then correctly identified as a token of malicious file information by the use of the contextual features. 

\subsection{Analysis of Contextual Features}
\label{sec:ana_context}

%%%%%%%%%%%%%%%%%%%
\begin{figure}[t]
\vspace{-0.2cm}
\centering
\includegraphics[scale=0.44]{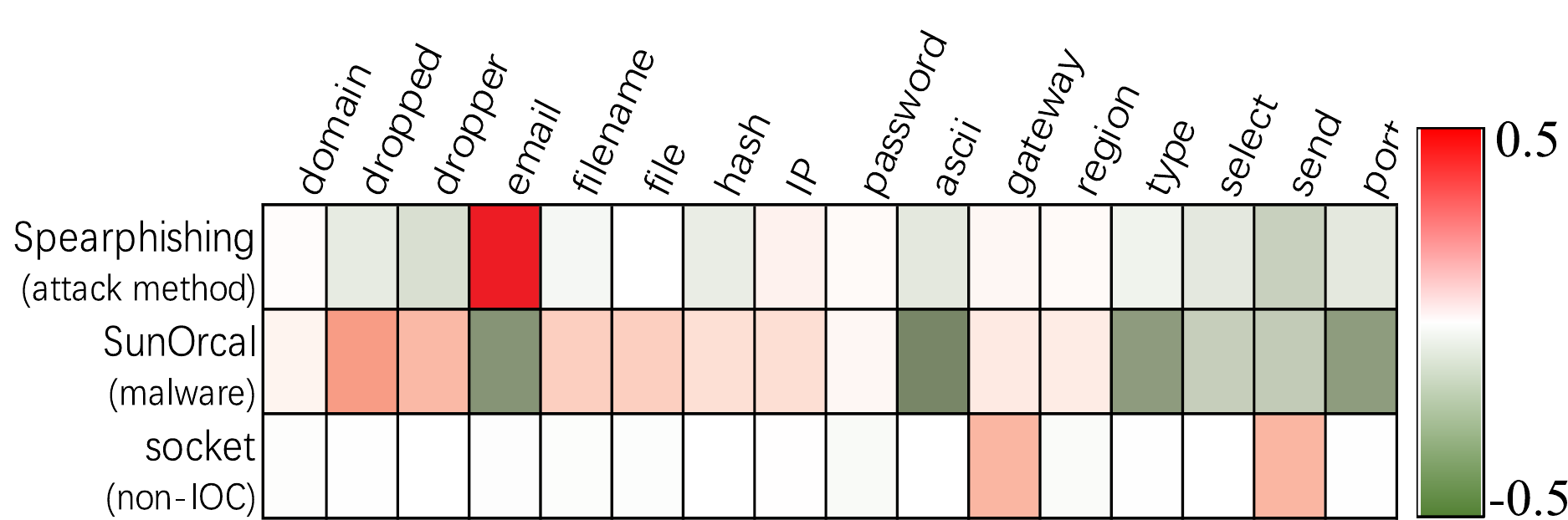}
\caption{Heatmap of part of contextual features martix in the English dataset}
\label{fig:ex_heatmap}
\end{figure}
%%%%%%%%%%%%%%%%%%

%%%%%%%%%%%%%%%%%%%
\begin{table}[t]
 \centering
 \caption{Top 10 and bottom 10 largest weighted contextual keywords of contextual feature in the English dataset}
 \label{tab:ex_context_keyword}
 \begin{tabular}{|p{1cm}||p{3cm}|p{3cm}|}
    \hline
     & top$-$10  & bottom$-$10 \\ \hline \hline
    IOC token & hash, domain, server, filename, ip, dropped, dropper, register, request, email & function, content, campaign, payload, key, sample, referred, specialized, sources, effort \\ \hline
    non-IOC token & ascii, password, researcher, eset, select, communicate, java, gateway, type, region & indicators, dropped, port, copies, lead, detection, dropper, send, PDB, register \\ \hline
  \end{tabular}
\end{table}
%%%%%%%%%%%%%%%%%%

 The proposed model provides an intuitive way to inspect the contextual information of each given token. As described in Section~\ref{sec:context_feature}, we initialize the contextual features of each given token using the automatically extracted contextual keywords and jointly learn them during the process of training with the whole ANN model.
 To prove the effectiveness of the contextual features, we visualize the learned weights martix of each contextual keyword of contextual feature and show several examples in Fig.~\ref{fig:ex_heatmap}. Each row of the matrix in each plot indicates the weights of contextual keywords for the given tokens. From this we see which contextual keyword were considered more important to represent the contextual information of the given token. 
 We can see from the matrix in Fig.~\ref{fig:ex_heatmap} that, for the token ``spearphshing'', which is an email-spoofing attack method, the contextual keyword ``email'' has the largest weight. For the malware ``SunOrcal'', which drops several malicious executable files, contextual keywords ``droppper'' and ``dropper'' have larger weights than other contextual keywords such as ``ascii'', ``port'' and ``type''. For non-IOC token ``socket'', contextual keywords ``gateway'' and ``port'' yield larger weights than other keywords because "socket" tends to co-occur with ``gateway'' and ``port''. 
 
 We further calculate the average weight of each contextual keyword and show the top 10 and bottom 10 largest weighted contextual keywords in TABLE~\ref{tab:ex_context_keyword}. From this we see that contextual keywords such as, ``hash'' and ``filename'', which tends to co-occur with malicious filenames, have the largest weights for IOCs, while the contextual keywords such as ``ascii'', ``password'' have the largest weights for non-IOCs. Here, it is interesting to find that contextual keyword ``dropped'' and ``droppper'', which tend to co-occur with malicious file information and malwares, yield large weights for IOCs but small weights for non-IOCs. 
 The proposed ANN model benefits from the differences of contextual information between IOCs and non-IOCs that is represented by the contextual features, and thus, achieves better performance than the previous works.

\subsection{Training the Proposed Model with Bilingual Data}
\label{sec:ana}

 Even though security articles are written in different languages, most of the IOCs are written in English, and are described in a similar pattern. Therefore, using multilingual corpora could be a solution for addressing the lack of annotated data, and the performance of the proposed model is expected to be improved by extending the training set. To examine the hypothesis, we ran a number of additional experiments using both the English dataset and Chinese dataset, both of which are described in Section~\ref{sec:data} and are not parallel data or comparable data.
 
%%%%%%%%%%%%%%%%%%%%%%%%%%%%%%%%%%%%%%%%%%%%%%
\begin{table}[t]
 \centering
 \caption{Comparison of evaluation results when training the proposed model with different training sets (micro average Precision / Recall / F1-score for 11 labels)}
 \label{tab:result_bi}
 \begin{tabular}{|c||c|c|}
    \hline
     & English test set & Chinese test set\\ \hline \hline
    Only English training set &\!\!\!{\bf 93.1} / 85.2 / 89.0\!\!\!&\!\!\!- / - / -\!\!\!\\ \hline
    Only Chinese training set &\!\!\!- / - / -\!\!\!&\!\!\!82.9 / 80.7 / 81.8\!\!\!\\ \hline \hline
    English $+$ Chinese training set &\!\!\!91.6 / {\bf 87.3} / {\bf 89.6}\!\!\!&\!\!\!{\bf 85.6} / {\bf 83.3} / {\bf 84.4}\!\!\!\\ \hline
  \end{tabular}
  \vspace{-0.2cm}
\end{table}
%%%%%%%%%%%%%%%%%%%%%%%%%%%%%%%%%%%%%%%%%%%%%

%%%%%%%%%%%%%%%%%%%%%%%%%%%%%%%%%%%%%%%%%%%
\begin{table*}[ht]
 \centering
 \caption{evaluation results for each label when training the proposed model with different training sets (Precision / Recall / F1-score) }
 \label{tab:label_result}
 \begin{tabular}{|p{3cm}||c|c||c|c|}
    \hline
     & \multicolumn{2}{c||}{English test set } & \multicolumn{2}{c|}{Chinese test set} \\ \cline{2-5}
     & Only English & English $+$ Chinese & only Chinese & English $+$ Chinese \\ 
     & training set & training set & training set & training set \\ \hline \hline
    attacker &  96.7 / 74.5 / 84.2 & 93.1 / 83.4 / {\bf 88.0} & 81.6 / 38.9 / 52.7 & 86.8 / 38.5 / {\bf 53.4} \\ \hline
    attack method & 90.6 / 93.4 / {\bf 92.0} & 90.7 / 91.9 / 91.3 & 16.3 / 29.6 / 21.1 & 70.0 / 33.7 / {\bf 45.5} \\ \hline
    attack target &  90.3 / 86.4 / {\bf 88.3} & 90.7 / 80.4 / 85.2 & 22.2 / 40.0 / 28.6 & 87.1 / 66.7 / {\bf 74.5} \\ \hline
    domain & 93.2 / 93.5 / 93.4 & 91.6 / 97.1 / {\bf 94.3} & 79.9 / 88.5 / 84.0 & 79.2 / 91.9 / {\bf 85.1} \\ \hline
    e-mail address & 98.4 / 84.4 / 90.8 & 97.5 / 91.0 / {\bf 94.1} & 20.4 / 20.4 / 20.4 & 81.3 / 53.1 / {\bf 64.2} \\ \hline
    file hash & 90.3 / 99.9 / {\bf 94.8} & 89.7 / 99.9 / 94.5 & 99.2 / 99.4 / {\bf 99.3} & 98.9 / 99.8 / {\bf 99.3} \\ \hline
    file information & 90.0 / 69.5 / 78.4 & 89.6 / 75.2 / {\bf 81.8}  & 58.0 / 87.2 / 67.7 & 60.9 / 88.0 / {\bf 74.1} \\ \hline
    IPv4 & 92.5 / 93.1 / 92.8 & 92.2 / 93.9 / {\bf 93.0} & 92.7 / 88.8 / {\bf 90.7} & 92.3 / 89.3 / {\bf 90.7}  \\ \hline
    malware & 97.3 / 68.8 / {\bf 80.6} & 96.4 / 67.9 / 79.6 & 93.9 / 59.7 / 69.7 & 92.6 / 64.0 / {\bf 72.0} \\ \hline
    URL & 99.7 / 92.3 / {\bf 95.9} & 98.9 / 92.5 / 95.6 & 82.5 / 83.1 / 82.8 & 84.9 / 86.2 / {\bf 85.5} \\ \hline
    vulnerability & 97.9  / 89.7 / 93.6 & 93.8 / 96.1 / {\bf 94.9} & 98.6 / 96.5 / {\bf 97.5} & 96.1 / 98.8 / {\bf 97.5} \\ \hline \hline
    micro average & 93.1 / 85.2 / 89.0 & 91.6 / 87.3 / {\bf 89.6} & 82.9 / 80.7 / 81.8 & 85.6 / 83.3 / {\bf 84.4} \\ \hline
 \end{tabular}
 \hspace{-0.2cm}
\end{table*}
%%%%%%%%%%%%%%%%%%%%%%%%%%%%%%%%%%%%%%%%%%%%

 As pre-trained word embeddings for the bilingual training dataset, we applied a cross-lingual word embedding obtained by the work of Duong el al\cite{Duong16}, where the English-Chinese cross-lingual dictionary is obtained by simply translating all the English words from English dataset to Chinese and Chinese words from Chinese dataset to English using Google translation\footnote{
  \url{https://translate.google.com}
 }.
 As contextual feature vector, we concatenate the contextual feature vector obtained from English dataset with the contextual feature vector obtained from Chinese dataset.
 Then we merge the English training set and the Chinese training set into one set and train the proposed model with the merged bilingual training set. 
%  {\color{red}We use only the English validation set when running experiments on the English test set and use only the Chinese validation set when running experiments on the Chinese test set.}
 TABLE~\ref{tab:result_bi} shows that the proposed model trained with the English training set and Chinese training set achieves a small improvement of F1-score on English test set when compared with the model trained with only English training set, and a great improvement of F1-score on Chinese test set when compared with the model trained with only Chinese training set.

 We compare scores of each label when the proposed model is trained with different training sets in TABLE~\ref{tab:label_result}. When using the English test set, the F1-scores of labels ``attack method'', ``attack target'' and ``malware'' by the model trained with the English training set and Chinese training set are lower than those scores by the model trained with only the English training set. It is mainly because that tokens of these labels can be written in different languages, which harms the model trained with the bilingual training data set. In contrast, benefiting from the extension of training set, for types of labels that are often written in English, e.g., ``domain '', ``file imformation'', ``IPv4'' and ``vlunerability'', the proposed model trained with the English training set and the Chinese training set achieves higher scores than the model trained with only the English training set. 
 When using the Chinese test set, the proposed model trained with the English training set and the Chinese training set obtained a obviously higher F1-scores than the model trained with only the Chinese training set for almost all the types of labels. It is interesting to find that types of labels ``e-mail address'', ``attack method'', ``attacker'', which lack of instances in Chinese training set, show the biggest improvement by using the model trained with the bilingual training set.  
 
%  Besides, we further ran experiments to evaluate the proposed model trained with only the English training set using the Chinese test set, and the proposed model trained with only Chinese training set suing the English test set. As shown in TABLE~\ref{tab:result_bi}, the average F1-score on the Chinese test set by the model trained with the English training set is 66.4\%, and the F1-score on the English test set by the model trained with the Chinese training set is 64.6\%, both of which are better than the scores of the baselines shown in TABLE~\ref{tab:result}.

\section{Related Work}
\label{sec:related}

\paragraph{NLP in cybersecurity}
Few references in cyber security utilize natural language processing. Neuhaus and Zimmermann\cite{Neuhaus10} analyze the trend of vulnerability by applying latent Dirichlet allocation to vulnerability description. Liao et al.\cite{LiaoX16} put forward a system to automatically extract IOC items from blog posts. Husari et al.\cite{Husari17} proposed a system that automatically extracted threat actions from unstructured threat intelligence reports by utilizing a pre-defined ontology. 
A concurrent work by Zhu et al.\cite{ZhuZ18} automatically extracted IOC data from security technical articles and further categorized them into different stages of malicious campaigns.
All of those systems consist of several components that rely heavily on manually defined rules, while our proposed model is an end-to-end model using word embedding and token features as input, which is more general and applicable to a broader area. 

\paragraph{Neural sequence labelling models}
There are amount of ANN-based works 
in the area of sequence labelling. Collobert et al.\cite{Collobert11} described one of the first task-independent neural sequence labelling model on the basis of convolutional neural networks. Hammerton\cite{Hammerton2003} first proposed a sequence labelling model with LSTM.  
Huang et al.~.\cite{HuangZ15} proposed using a sequence labelling model based on the bidirectional LSTM for the task of name entity recognition (NER). Lample et al.\cite{Lample16} proposed integrating LSTM encoders with character embedding and the neural sequence labelling model.
Rei et al.\cite{ReiM16} improved the model by introducing an attention mechanism to the character-level representations.
Dernoncourt et al.\cite{Dernoncourt17} proposed applying the neural sequence labelling model to the task of de-identification of medical records. 
Liu et al.\cite{LiuL18} proposed a sequence labeling framework, which effectively leverages the language model to extract character-level knowledge.
One appealing property of those works is that they can achieve excellent performance with a unified architecture and without task-specific feature engineering. It remains unclear whether such works can be used for tasks without a large amount of training data. Several works such as Yang et al.\cite{YangZ17} and Lee et al.\cite{LeeJY18} proposed applying transfer learning to NER using a limited number of training corpora. Nevertheless, a large dataset that has same labels as the small training dataset is required for transfer learning, which is hard to obtain in the field of cybersecurity. In this paper, we introduce several spelling features without expert knowledge of cybersecurity to the neural model, and achieve excellent performance even using a small dataset for training.

\section{Conclusions}
\label{sec:conc}
 To conclude, in this paper, we newly introduce a multi-head self-attention module and contextual features to the neural based sequence labelling model, which significantly improved the performance in the task of IOC identification.
%  and evaluate the proposed model on English APT reports and Chinese cybersecurity blog articles.
 Based on the evaluation results of our experiments, our proposed model 
 is proved effective on both the English test set and the Chinese test set.
% %  approach
%  performs better than other sequence labelling models with an average F1-score of 89.0\% on English test set and an average F1-score of 81.8\% on Chinese test set.
 We further evaluated the proposed model by training the proposed model using both the English training set and the Chinese training set and compared it with models that are trained with only one training set, where the model trained with the merged bilngual training set performs better. 
 
 One of our future works is to integrate the contextual embeddings from the bidirectional language model into our proposed model. The pretrained neural language models are proved effective in the sequence labelling models\cite{Peter17,Xiong18,Peters18}. 
%  However, our preliminary experiments show that sequence labelling models with neural language model require a large amount of training data and are inappropriate for the task of IOCs identification. Therefore, 
 It is expected to improve the performance of the proposed model by integrating both the contextual features and contextual embeddings into the neural sequence labelling model. 

% ------------------
\bibliographystyle{IEEEtran}
\bibliography{long}

% Generated by IEEEtran.bst, version: 1.14 (2015/08/26)
\begin{thebibliography}{10}
\providecommand{\url}[1]{#1}
\csname url@samestyle\endcsname
\providecommand{\newblock}{\relax}
\providecommand{\bibinfo}[2]{#2}
\providecommand{\BIBentrySTDinterwordspacing}{\spaceskip=0pt\relax}
\providecommand{\BIBentryALTinterwordstretchfactor}{4}
\providecommand{\BIBentryALTinterwordspacing}{\spaceskip=\fontdimen2\font plus
\BIBentryALTinterwordstretchfactor\fontdimen3\font minus
  \fontdimen4\font\relax}
\providecommand{\BIBforeignlanguage}[2]{{%
\expandafter\ifx\csname l@#1\endcsname\relax
\typeout{** WARNING: IEEEtran.bst: No hyphenation pattern has been}%
\typeout{** loaded for the language `#1'. Using the pattern for}%
\typeout{** the default language instead.}%
\else
\language=\csname l@#1\endcsname
\fi
#2}}
\providecommand{\BIBdecl}{\relax}
\BIBdecl

\bibitem{ZhuZ16}
Z.~Zhu and T.~Dumitras, ``Feature{S}mith: automatically engineering features
  for malware detection by mining the security literature,'' in \emph{Proc. CCS
  16}, 2016, pp. 767--778.

\bibitem{LiaoX16}
X.~Liao, K.~Yuan, X.~Wang, Z.~Li, L.~Xing, and R.~Beyah, ``Acing the ioc game:
  toward automatic discovery and analysis of open-source cyber threat
  intelligence,'' in \emph{Proc. CCS 16}, 2016, pp. 755--766.

\bibitem{Husari17}
G.~Husari, E.~Al-Shaer, M.~Ahmed, B.~Chu, and X.~Niu, ``{TTPD}rill: automatic
  and accurate extraction of threat actions from unstructured text of cti
  sources,'' in \emph{Proc. ACSAC 2017}, 2017, pp. 103--112.

\bibitem{HuangC17}
C.~Huang, S.~Hao, L.~Invernizzi, J.~Liu, Y.~Fang, C.~Kruegel, and G.~Vigna,
  ``Gossip: automatically identifying malicious domains from mailing list
  discussions,'' in \emph{Proc. ASIA CCS 17}, 2017, pp. 494--505.

\bibitem{Kwon17}
B.~J. Kwon, V.~Srinivas, A.~Deshpande, and T.~Dumitras, ``Catching worms,
  trojan horse and {PUP}s: unsupervised detection of silent delivery
  campagins,'' in \emph{Proc. NDSS 17}, 2017.

\bibitem{ZhuZ18}
Z.~Zhu and T.~Dumitras, ``Chain{S}mith: automatically learning the semantics of
  malicious campaigns by mining threat intelligence reports,'' in \emph{Proc.
  EuroS\&P 2018}, 2018.

\bibitem{HuangZ15}
Z.~Huang, W.~Xu, and K.~Yu, ``Bidirectional lstm-crf models for sequence
  tagging,'' \emph{arXiv:1508.01991}, 2015.

\bibitem{Hochreiter97}
S.~Hochreiter and J.~Schmidhuber, ``Long short-term memory,'' \emph{Neural
  computation}, vol.~9, no.~8, pp. 1735--1780, 1997.

\bibitem{Chiu16}
J.~P. Chiu and E.~Nichols, ``Named entity recognition with bidirectional
  {LSTM}-{CNN}s,'' \emph{Transactions of the Association for Computational
  Linguistics}, vol.~4, pp. 357--370, 2016.

\bibitem{Lample16}
G.~Lample, M.~Ballesteros, S.~Subramanian, K.~Kwakami, and C.~Dyer, ``Neural
  architectures for name entity recognition,'' in \emph{Proc. NAACL 2016},
  2016, pp. 260--270.

\bibitem{Dernoncourt17}
F.~Dernoncourt, J.~Y. Lee, O.~Uzuner, and P.~Szolovits, ``De-identification of
  patient notes with recurrent neural networks,'' \emph{Journal of the American
  Medical Informatics Association}, vol.~24, no.~3, pp. 596--606, 2017.

\bibitem{JiangZ17}
Z.~Jiang, C.~Zhao, B.~He, Y.~Guan, and J.~Jiang, ``De-identification of medical
  records using conditional random fields and long short-term memory
  networks,'' \emph{Journal of Biomedical Informatics}, vol.~75, pp. S43--S53,
  2017.

\bibitem{Zhou18}
S.~Zhou, Z.~Long, L.~Tan, and H.~Guo, ``Automatic identification of indicators
  of compromise using neural-based sequence labelling,'' in \emph{Proc. PACLIC
  2018}, 2018.

\bibitem{Yang16}
Z.~Yang, D.~Yang, C.~Dyer, X.~He, A.~Smola, and E.~Hovy, ``Hierarchical
  attention networks for document classification,'' in \emph{Proc. NAACL-HLT
  2016}, 2016, pp. 1480--1489.

\bibitem{Vaswani17}
A.~Vaswani, N.~Shazeer, N.~Parmar, J.~Uszkoreit, L.~Jones, A.~N. Gomez, L.~.
  Kaiser, and I.~Polosukhin, ``Attention is all you need,'' in \emph{Proc. 31st
  NeurIPS}, 2017, pp. 5998--6008.

\bibitem{XieJ18}
J.~Xie, Z.~Yang, G.~Neubig, N.~A. Smith, and J.~Carbonell, ``Neural
  cross-lingual named entity recognition with minimal resources,'' in
  \emph{Proc. EMNLP 2018}, 2018, pp. 369--379.

\bibitem{Darling15}
M.~Darling, G.~Heileman, G.~Gressel, A.~Ashok, and P.~Poornachandran, ``Lexical
  approach for classifying malicious urls,'' in \emph{IEEE International
  Conference on High Performance Computing \& Simulation (HPCS)}, 2015, pp.
  195--202.

\bibitem{Mikolov13}
T.~Mikolov, I.~Sutskever, K.~Chen, G.~S. Corrado, and J.~Dean, ``Distributed
  representations of words and phrases and their compositionality,'' in
  \emph{Advances in neural information processing systems}, 2013, pp.
  3111--3119.

\bibitem{ReiM16}
M.~Rei, G.~K.~O. Crichton, and S.~Pyysalo, ``Attending to characters in neural
  sequence labeling models,'' in \emph{Proc. COLING 2016}, 2016.

\bibitem{Duong16}
L.~Duong, H.~Kanayama, T.~Ma, S.~Bird, and T.~Cohn, ``Learning crosslingual
  word embeddings without bilingual corpora,'' in \emph{Proc. EMNLP 2016},
  2016, pp. 1285--1295.

\bibitem{Neuhaus10}
S.~Neuhaus and T.~Zimmermann, ``Security trend analysis with {CVE} topic
  model,'' in \emph{IEEE 21st International Symposium on Software Reliability
  Engineering}, 2010, pp. 111--120.

\bibitem{Collobert11}
R.~Collobert, J.~Weston, L.~Bottou, M.~Karlen, K.~Kavukcuoglu, and P.~Kuksa,
  ``Natural language processing (almost) from scatch,'' \emph{The Journal of
  Machine Learning Research}, vol.~12, pp. 2493--2537, 2001.

\bibitem{Hammerton2003}
J.~Hammerton, ``Named entity recognition with long short-term memory,'' in
  \emph{Proc. CONLL 2003}, 2003, pp. 172--175.

\bibitem{LiuL18}
L.~Liu, J.~Shang, X.~Ren, F.~F. Xu, H.~Gui, J.~Peng, and J.~Han, ``Empower
  sequence labeling with task-aware neural language model,'' in \emph{Proc.
  32nd AAAI}, 2018, pp. 5253--5260.

\bibitem{YangZ17}
Z.~Yang, R.~Salakhutdinov, and W.~W. Cohen, ``Transfer learning for sequence
  tagging with hierarchical recurrent networks,'' in \emph{Proc. ICLR 2017},
  2017.

\bibitem{LeeJY18}
J.~Y. Lee, F.~Dernoncourt, and P.~Szolovits, ``Transfer learning for
  named-entity recognition with neural networks,'' in \emph{Proc. LERC 2018},
  2018, pp. 4471--4473.

\bibitem{Peter17}
M.~E. Peters, W.~Ammar, C.~Bhagavatula, and R.~Power, ``Semi-supervised
  sequence tagging with bidirectional language models,'' in \emph{Proc. 55th
  ACL}, 2017, pp. 1756--1765.

\bibitem{Xiong18}
Y.~Xiong and M.~Nuo, ``Attention-based blstm-crf architecture for mongolian
  named entity recognition,'' in \emph{Proc. PACLIC 2018}, 2018.

\bibitem{Peters18}
M.~E. Peters, M.~Neumanm, M.~Iyyer, and M.~Gardner, ``Deep contextualized word
  representations,'' in \emph{Proc. NAACL-HLT 2018}, 2018, pp. 2227--2237.

\end{thebibliography}
%------------------

\end{document}